%% file: main.tex
\documentclass[a4paper, twocolumn]{article} 

\usepackage{latexsym}
\usepackage{amssymb}
\usepackage{amsmath}
\usepackage{amsthm}
\usepackage{booktabs}
\usepackage{enumitem}
\usepackage{graphicx}
\usepackage{color}
\usepackage{authblk}
\usepackage[numbers]{natbib}
\usepackage{url}
\usepackage{textcomp}
\usepackage{hyperref}
\usepackage[font=footnotesize,skip=2pt]{caption}
\usepackage{adjustbox}

\begin{document}

\title{Prompt-Based Value Steering of Large Language Models }
\author[]{Giulio Antonio Abbo}
\author[]{Tony Belpaeme}

\affil[]{\textit{IDLab-AIRO}, \textit{Ghent University -- imec}, Ghent, Belgium}
\affil[]{\texttt{giulioantonio.abbo@ugent.be}}

\date{\scriptsize Presented at the 3rd International Workshop on Value Engineering in AI (VALE 2025), 28th European Conference on AI.\\To appear in Springer LNCS.}

\maketitle


\begin{abstract}
Large language models are increasingly used in applications where alignment with human values is critical. While model fine-tuning is often employed to ensure safe responses, this technique is static and does not lend itself to everyday situations involving dynamic values and preferences. In this paper, we present a practical, reproducible, and model-agnostic procedure to evaluate whether a prompt candidate can effectively steer generated text toward specific human values, formalising a scoring method to quantify the presence and gain of target values in generated responses. We apply our method to a variant of the Wizard-Vicuna language model, using Schwartz's theory of basic human values and a structured evaluation through a dialogue dataset. With this setup, we compare a baseline prompt to one explicitly conditioned on values, and show that value steering is possible even without altering the model or dynamically optimising prompts.
\end{abstract}



\section{Introduction}

Large Language Models are widely used to generate text for the most diverse uses, including conversation~\cite{app15116377}, summarisation, and storytelling~\cite{10974188}.
In these tasks, it is important to ensure that the output produced by LLMs reflects users' preferences and desirable human values.
However, these concepts can be abstract and complex, and ensuring that generated content aligns with them is a nontrivial challenge.

Many studies have explored ways to guide LLMs through either prompt engineering~\cite{li2025survey} or direct intervention on the model itself, such as fine-tuning or reinforcement learning~\cite{bai2022training}.
The latter is the preferred choice when it comes to the value alignment and safety of these tools: indeed, many models implicitly favour certain values over others~\cite{yao2025value}.

Having these values frozen and embedded in the model weights becomes a problem in situations where the model's behaviour should adapt dynamically to the conversation participants, keeping into consideration their preferences, the context, and the social norms involved.
Furthermore, in some instances changing the structure or fine-tuning a model is not feasible.
The solution is to control the alignment of the responses through the prompt, the instructions given to the LLM.

In this work, we explore how far value alignment can be achieved through prompt design alone, without modifying the model's parameters.
To this end, we propose a procedure to evaluate the effectiveness of a prompt in producing value-aligned responses.
We demonstrate the application of our method, combining a value classifier (ValuesNet\_DeBERTa\_v3), a structured theory of values (Schwartz's model), and a dialogue dataset (Commonsense-Dialogues) to measure the presence of values in generated outputs.

Our method assumes no changes to the model or prompt beyond those externally defined, and quantifies alignment in a systematic manner.
While we focus on a specific value theory and domain, the approach is general and can be adapted to other taxonomies or use cases.
By choosing the values and datasets that most fit a specific application, through this procedure it is possible to obtain insights into how certain prompts will perform with certain language models in that application.
We see this as an important step toward better understanding and controlling how dynamic values are expressed in language model outputs.


\section{Related work}

Human values have been studied and classified in established psychological theories.
Among these, Schwartz's theory of basic human values~\cite{schwartz2012overview} has been used to define interpretable and culturally universal value categories.
In its simpler formulation, it comprises ten values that represent broad motivational goals: \textit{benevolence}, \textit{universalism}, \textit{self-direction}, \textit{stimulation}, \textit{hedonism}, \textit{achievement}, \textit{power}, \textit{security}, \textit{conformity}, and \textit{tradition}.
These values are organised in a circular structure reflecting compatibilities and conflicts.
Alternative frameworks include Moral Foundations Theory~\cite{graham2013moral} and Rokeach's value survey~\cite{rokeach1967rokeach}.

Recent work has focused on annotating datasets with values, an example are the Touché23-ValueEval dataset~\cite{mirzakhmedova-etal-2024-touche23} and the ValueNet dataset~\cite{Qiu_Zhao_Li_Lu_Peng_Gao_Zhu_2022}.
These provide large-scale value annotations for social media and dialogue data, enabling supervised training and evaluation of models such as DeBERTa~\cite{he2023debertav} for automatic value classification.
In practice, given a short text, these models can determine its alignment with each of the values of a theory.
Competitions and events such as Touché~\cite{touche24_proceedings} allow evaluating and comparing the performances of these classifiers.

Beyond value classification, recent work has also investigated how well large language models align with human values in their generated outputs.
Several studies have proposed evaluation benchmarks and assessed whether models exhibit desirable traits by analysing their responses to moral dilemmas and judgement tasks~\cite{10.1145/3715275.3732044,NEURIPS2023_f751c6f8}, surveys~\cite{NEURIPS2023_a2cf225b}, scenarios~\cite{10.1007/978-3-031-58202-8_6} and images~\cite{10.1007/978-3-031-85463-7_5}.
Moving away from fixed questionnaires, \citet{yao2025value} propose a generative paradigm to assess the value alignment of LLMs, by examining the value conformity reflected in open-ended responses through a value recogniser.
These efforts typically treat alignment as a fixed property of a model, resulting from its training or fine-tuning.
In contrast, our work shifts the focus to the prompt as the source of value alignment and investigates whether prompts alone can reliably elicit outputs aligned with specific human values.

Traditionally, LLMs' output control is achieved either through reinforcement learning~\cite{bai2022training} and techniques that change the model structure or weights, or through prompt design; in this work, we are interested in the latter.
Prompt-based control of language generation, without modifying model weights, has received significant attention, especially through the use of contextual information, goals, and common patterns~\cite{white2023prompt}.
While initially used for task adaptation through few-shot learning -- when a model trained for one task is used for another, similar task, by providing some examples of the desired output -- prompt crafting has increasingly been used to align generation with stylistic or goal-driven constraints, thanks to the advantage of not having to fine-tune the language model, saving time and costs.

Finding the most performant prompt structure by hand, however, is not easy.
Prompt engineering is the process of crafting input prompts, testing them, and improving them iteratively.
Several works have proposed automated or semi-automated methods to search for effective prompts~\cite{li2025survey}, leveraging techniques such as gradient-guided search~\cite{shin-etal-2020-autoprompt}, genetic algorithms~\cite{secheresse2025gaapo}, or the use of LLMs to improve the prompt itself~\cite{ye-etal-2024-prompt}.
These methods rely on modifying the prompt automatically, following a measure to be optimised.
Our work falls under this category, as we propose a score to determine the ability of a prompt to steer the generated output towards certain values.
Consequently, this score can be used as a target metric for other prompt iteration and improvement algorithms.


\section{Evaluation procedure}

\begin{table}[t]
\caption{Procedure results with control variables and final score.}
\label{tab:variables}
\centering
\begin{adjustbox}{width=\linewidth,center=\linewidth}
\input{tab-variables}
\end{adjustbox}
\end{table}

The objective of the procedure is to evaluate how well a prompt candidate can steer an LLM's output towards maximising any chosen value of a given value theory.
The process depends on four components: the target LLM, a value detector, a dataset of test inputs, and a prompt candidate; these are described in detail in Section~\ref{sec:control-variables}.
The procedure for the score calculation, reported in Section~\ref{sec:score-calculation}, involves four steps.
1)~The value detector extracts the initial values from each element of the dataset of test inputs.
2)~Each element is then combined with the prompt candidate and fed to the target LLM, generating a response that should maximise -- one by one -- each of the values under consideration.
3)~The value detector extracts the response values from each of the generated completions.
4)~An analysis of the initial and final values of each completion allows computing the prompt score.
At this point, the procedure can be repeated with a new prompt candidate, which can be crafted manually or with any automated technique.
Finally, the best-performing prompt can be tested again, on a split of the original dataset or on real data.

\subsection{Ensuring experimental consistency}
\label{sec:control-variables}

The reliability of the score depends on a clear description of any fixed aspect that has an effect on the results.
We propose the following control variables, divided into four groups, which can be expanded to accommodate the needs of each application using this procedure.
This information can be presented in a compact format and complements the procedure results as in Table~\ref{tab:variables}.

The first group concerns the \textbf{target LLM}.
This is the model that will be used to generate responses based on the candidate prompt.
For reproducibility, it is important that the variant or snapshot of the model used is specified, together with the LLM's parameters, such as the temperature, if they differ from the defaults of the model.

A second fundamental aspect is the value theory used.
Once the value theory has been established, it is necessary to specify which \textbf{value detector} is used to extract the values from the text.
The parameters used for the chosen method should include details such as thresholds for label acceptance and how the input is analysed -- e.g., the whole input or only the last 200 characters.
Since each theory can be applied with slight variations, the values used should be reported in a value list.
For instance, some value detectors indicate only the presence of a value, while others distinguish between alignment, neutrality, and opposition to a value (e.g., classifying text as expressing benevolence, being neutral, or expressing malevolence).
If possible, the value detector should be validated, to obtain an idea of its implementation's performance.

The third element is the \textbf{dataset of test inputs} used to generate the responses with the prompt candidate.
Relevant aspects in this area are the dataset structure -- e.g., social media posts, forum entries, or dialogues -- and the dataset split, specifying the percentage of data points used to run the procedure and potentially to evaluate it, including any information useful for the reproducibility of the experiment.

A section with the prompt scores completes the results report.
In particular, we suggest reporting both the scores of the \textbf{prompt candidate} and a baseline prompt, which is not designed to favour any specific value.
This information is useful as every model tends to give responses which are aligned with some values by default~\cite{yao2025value}.

\subsection{Score calculation}
\label{sec:score-calculation}

We formalise the score calculation using the following equation.
Given $V$ the set of values under consideration and $X$ the set of all possible strings, we call $Eval_v$ a boolean function representing the value detector.
$Eval_v(x)$ returns 1 if the string $x \in X$ shows a value $v \in V$, $0$ otherwise; its negation, $\overline{Eval}$, behaves in the opposite way, returning $1$ if the string does not display the value $v$.
If we name $D \subset X$ the dataset of test inputs, we can define the function $C~:~(v,d)~\mapsto~x$ as the prompt candidate.
This function combines into a string $x$ the instructions to maximise a value $v$ with the test input $d \in D$.
For each value $v$ we define the following functions.

\begin{align}
Gains_v = \sum_{d \in D} \quad \overline{Eval_v}(d) \wedge {Eval_v}(C(v,d))     \\
Retains_v = \sum_{d \in D} \quad {Eval_v}(d) \wedge {Eval_v}(C(v,d))   \\
Losses_v = \sum_{d \in D} \quad {Eval_v}(d) \wedge \overline{Eval_v}(C(v,d))    \\
Neutrals_v = \sum_{d \in D} \quad \overline{Eval_v}(d) \wedge \overline{Eval_v}(C(v,d))
\end{align}

$Gains_v$ represents the number of instances in the dataset where the value $v$ was not in the initial input, and was added by the prompt candidate $C$. Similarly, $Retains_v$ counts the times a value is present both before and after the prompt candidate is applied, $Losses_v$ 
considers the case where the value is lost by applying the prompt candidate, and $Neutrals_v$ where it is undetected both before and after.
Splitting the evaluation into multiple measures allows for accounting for the initial values distribution in the test dataset.
We combine these measures as follows to obtain a prompt score $S_v$ for each value.

\begin{equation}
S_v = \alpha Gains_v + \beta Retains_v + \gamma Losses_v + \delta Neutrals_v
\end{equation}

The coefficients are manually set and allow controlling the importance of each measure; we consider the first two to be positive and the last two negative.
Since gaining a value or retaining it are equally successful outcomes, we assign $\alpha = \beta = 1$.
To penalise a value loss when maximising for that value, we assign $\gamma = -1$.
Finally, we set $\delta = -0.5$, to penalise those cases where a string remains value-neutral, but with less importance than a loss, since there are situations where a value is not applicable, especially with short text generations.
We then obtain $\hat S_v$ by normalising the score over the interval $[0,1]$ with the following equations.

\begin{align}
S_{v,max} &= \sum_{d \in D} \alpha \overline{Eval_v}(d) + \beta {Eval_v}(d)\\
S_{v,min} &= \sum_{d \in D} \gamma {Eval_v}(d) + \delta \overline{Eval_v}(d)\\
\hat S_v &= \frac{S_v - S_{v,min}}{S_{v,max} - S_{v,min}}
\end{align}

Finally, we can compute the final score $S$ which encompasses all values.
If necessary, it is possible to give different weights to the values, if in a certain application some are more important than others.
This should be reported in the procedure results table.
We will consider all values to be equally important, thus $S$ is defined as the arithmetic mean of all $S_v$.


\section{Case study}

\begin{table}[t]
\caption{Value detector validation scores. The accuracy, F1 macro and F1 weighted scores take into account that each value label can have positive, negative or neutral valence. The final weighted mean takes into account the support of each value in the dataset.}
\label{tab:validation}
\centering
\begin{adjustbox}{width=\linewidth,center=\linewidth}
\input{tab-validation}
\end{adjustbox}
\end{table}

\begin{table*}[t]
\caption{Intermediate results for the baseline and candidate prompt. $E_v$ represents the count of instances in the dataset aligned with the value $v$ and $\bar E_v$ those neutral or against it. $S_{v,min}$ is the minimum possible score for each value; the maximum is always 1K due to the coefficients chosen and is omitted. $G_v$, $R_v$, $L_v$, $N_v$ are the number of instances where the value $v$ was gained, retained, lost, or not acquired when trying to maximise for $v$. $S_v$ and $\hat S_v$ are the score and normalised scores for each value (higher is better). The difference between the two $\hat S_v$ is reported under $\Delta \hat S_v$.}
\label{tab:intermediate}
\centering
\begin{adjustbox}{width=\textwidth,center=\textwidth}
\input{tab-intermediate}
\end{adjustbox}
\end{table*}

For a case study of how to execute the procedure in practice, we select the values from Schwartz's basic human values theory.
As the value detector, we use Valuesnet\_DeBERTa\_v3~\cite{nick_h_2024,Qiu_Zhao_Li_Lu_Peng_Gao_Zhu_2022} replicating the setup of the model's author.
This model takes a sentence as input and a value from Schwartz's theory, returning $1$ if the string is aligned with the value, $-1$ if it is not aligned, $0$ if it is neutral.
We validate our results on the test set of the ValueNet Dataset~\cite{Qiu_Zhao_Li_Lu_Peng_Gao_Zhu_2022} obtaining an F1 score of 0.66 as reported in Table~\ref{tab:validation}.
We consider a value to be present if the value detector finds alignment and we treat both sentences that oppose a value and those that are neutral as lacking that value.
Other applications can introduce new values: for instance, \textit{malevolence} could be used to represent misalignment with \textit{benevolence}.
We evaluated other classifiers for this application and decided to exclude them due to insufficient information on how to reimplement the authors' setup or due to low F1 scores on the test set.

\begin{figure*}[t]
\centering
\includegraphics[width=0.8\linewidth]{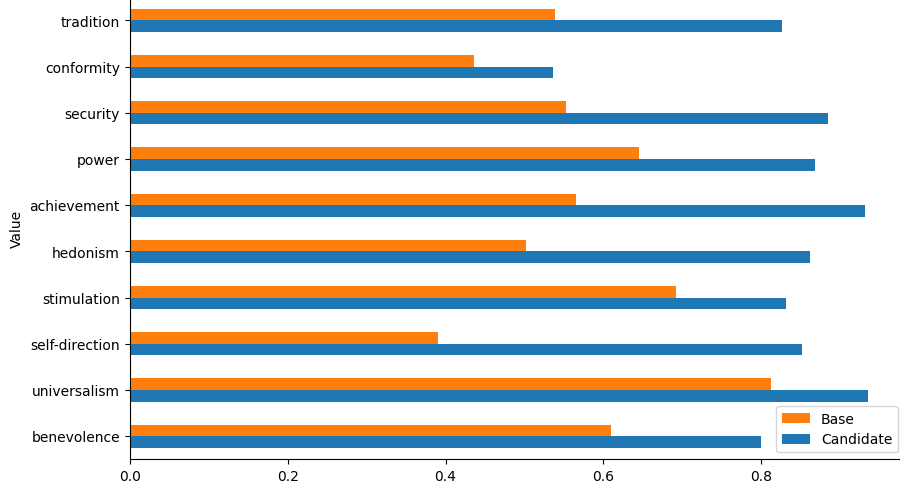}
\caption{Comparison of the normalised scores for the baseline and candidate prompts, across values.}
\label{fig:comparison}
\end{figure*}

\begin{table}[ht!]
\caption{Procedure results presentation for the case study.}
\label{tab:results}
\centering
\small
\input{tab-results}
\end{table}

To test the prompt candidate, we use 1K samples from the train split of the Commonsense-Dialogues dataset~\cite{zhou-etal-2021-commonsense}.
When generating the dialogue continuation using the candidate prompt, for each conversation all the dialogue turns are used.
Instead, when extracting the values from each dialogue, only the last two turns are used, or the last turn and the generated text.
This precaution increases the weight of the values in the generated text while at the same time ensuring that the evaluated string contains enough contextual information.

The target LLM is a self hosted instance of Wizard-Vicuna-13B-Uncensored\footnote{\url{https://huggingface.co/TheBloke/Wizard-Vicuna-13B-Uncensored-GGUF}} obtained by excluding the moral guardrails\footnote{\url{https://huggingface.co/cognitivecomputations/Wizard-Vicuna-13B-Uncensored}} from a model inspired\footnote{\url{https://github.com/melodysdreamj/WizardVicunaLM}} by WizardLM~\cite{xu2024wizardlm} and VicunaLM~\cite{NEURIPS2023_91f18a12}.
We use an uncensored model as alignment guardrails might interfere with our setup.
We use a temperature of $0$ to improve repeatability and leave the other settings to their defaults\footnote{\url{https://github.com/abetlen/llama-cpp-python}}.
The prompt template used is the same as the Vicuna models, reported here with an example dialogue, where \textlangle{}SYSTEM\textrangle{} and \textlangle{}COMMAND\textrangle{} are placeholders that depend on the prompt candidate.

\begin{quote}
\textlangle{}SYSTEM\textrangle{}\\
USER: I know what she was going through was hard but this is just a temporary feeling.\\
ASSISTANT: And she is young indeed.\\
USER: Everything will be okay. I'll be there for her.\\
ASSISTANT: It must hurt so much now though.\\
USER: I told her that after every bad thing that happens, there is good that happens. She is a good person and good things will happen to her.\\
ASSISTANT: Kai that was really thoughtful of you.
\\
USER: \textlangle{}COMMAND\textrangle{}\\
ASSISTANT:
\end{quote}

Following our procedure, we extract the initial values, generate the responses for each dialogue and each value, and calculate the score for each value and a final score.
We test a baseline prompt with system message \textit{``You are having a conversation.''} and the command \textit{``Generate a short response.''}
Then we test a prompt candidate with the system message \textit{``You are having a conversation, your responses are based on a specific value.''} and the command \textit{``Generate a short response that aligns with the value `\textlangle{}VALUE\textrangle{}'.''} where the value is inserted at runtime.

For illustration purposes, we report the intermediate results of both prompts in Table~\ref{tab:intermediate}.
Due to $\alpha$ and $\beta$ being equal to $1$, $S_{v,max}$ is always equal to the number of samples, while $S_{v,min}$ depends on how many samples were initially aligned with each value.
Comparing the score $\hat S_v$ of each value in the baseline case, we observe that the model is not perfectly neutral: it exhibits a slight tendency to express or increase most values by default, particularly \textit{universalism} and \textit{stimulation}.
In contrast, values like \textit{self-direction} and \textit{conformity} show weaker spontaneous presence, suggesting the model's underlying behaviour is biased toward certain value profiles.
The candidate prompt improved not only value retention in already-aligned cases, but also increased value gains.
Meanwhile, losses and neutrals decreased substantially.
Certain values, such as \textit{universalism}, \textit{achievement}, and \textit{security}, benefited more from the candidate prompt.
This may be due to their more concrete or action-oriented nature, which is easier to express in short conversational turns.
In contrast, values like \textit{conformity} showed less improvement, possibly due to their abstract or socially contextual nature.
If we compare the scores of the two prompts -- $\Delta \hat S_v$ in Table~\ref{tab:intermediate}, and Figure~\ref{fig:comparison} -- we see an improvement across all values.

The results demonstrate that conditioning the LLM on explicit values significantly improves alignment.
The final score is reported in Table~\ref{tab:results}, where it is evident that the prompt performance improved from $0.57$ in the baseline to $0.83$ with the candidate prompt, confirming that explicitly stating the value resulted in more consistent value-aligned behaviour.


\section{Conclusion}

We propose a score to compare the success rate of given prompt candidates to steer the output of an LLM towards a chosen component of a value theory.
This measure takes into account the values present in an initial text from a dataset and those in its LLM-generated continuations using the prompt candidate.
The method differentiates between values gained, values kept, lost and neutral sentences, thus remaining unbiased by the initial value distribution of the dataset.
The score can be used in an iterative prompt engineering procedure to obtain a best-performing prompt.
The procedure results are dependent on a series of variables; to ensure the results maintain their relevance in practical applications, we propose a format to report the results together with these variables. The procedure is available online\footnote{\url{https://github.com/giubots/value-steering}, archived at \url{https://doi.org/10.5281/zenodo.17609013}}.

While the results are promising, limitations remain.
The analysis focuses on a single model, and the evaluation method treats neutral and misaligned outputs equally.
Further work could treat misalignment with a value as a separate label, allowing to measure if the prompts can maximise the opposite of each value.
It is also possible to study the effect of maximising one value across all the others, possibly using numerical data for the labels instead of categorical ones.
Additionally, exploring multi-turn conditioning and stylistic variations may help balance value alignment with conversational naturalness.
Our results depend strongly on the performance of the value classifier: we accommodate for this by including all relevant info in the results table, improving their reproducibility.
However, future research should explore more reliable value extraction techniques.
Finally, in our study we did not consider variables such as response time, which play a fundamental role when using long prompts in real-time scenarios.

With this research, we aim to advance the field of value-aligned prompt engineering by offering a clear structure for both evaluation and reporting.
Beyond its immediate application, the proposed framework can support broader investigations into how prompts interact with model behaviours across contexts, value systems, and input types. It also opens opportunities for integrating this approach into automated alignment pipelines, comparative benchmarking, or real-time deployment scenarios where sensitivity to dynamic values is critical.


\section*{Acknowledgments}
Funded by the Horizon Europe VALAWAI project (grant agreement number 101070930).


\bibliographystyle{plainnat}
\bibliography{bibliography}
\end{document}

%% file: tab-variables.tex
\begin{tabular}{@{}ll@{}}
\toprule
Variable               & Description                                                   \\
\midrule
\midrule
Target LLM name        & Name and version of the target model.                         \\
Target LLM parameters  & Parameters modified of the target model.                      \\
\midrule
Value theory           & Name of the theory used.                                      \\
Value list             & List of values used in the theory.                            \\
Method name            & Method used to extract the values.                            \\
Method parameters      & Parameters used in the method.                                \\
\midrule
Dataset name           & Name of the dataset used for testing.                         \\
Dataset type           & Type of dataset (e.g., tweets, ...)                           \\
Dataset split          & Percentages of the dataset used (if applicable).              \\
\midrule
Score p. baseline      & Obtained with a generic prompt.                               \\
Score p. candidate     & Obtained with the prompt candidate.                           \\
\bottomrule
\end{tabular}

%% file: tab-validation.tex
\begin{tabular}{@{}lrrr@{}}
\toprule
Value          & Accuracy & F1 macro & F1 weighted \\
\midrule
Benevolence    & 0.70     & 0.63     & 0.70        \\
Universalism   & 0.58     & 0.52     & 0.58        \\
Self-direction & 0.52     & 0.44     & 0.50        \\
Stimulation    & 0.63     & 0.63     & 0.63        \\
Hedonism       & 0.71     & 0.71     & 0.71        \\
Achievement    & 0.64     & 0.65     & 0.64        \\
Power          & 0.55     & 0.47     & 0.54        \\
Security       & 0.51     & 0.50     & 0.51        \\
Conformity     & 0.82     & 0.57     & 0.81        \\
Tradition      & 0.73     & 0.67     & 0.73        \\
\multicolumn{3}{l}{\textbf{Weighted mean}}    & \textbf{0.66 }       \\
\bottomrule
\end{tabular}

%% file: tab-intermediate.tex
\begin{tabular}{@{}lrrr|rrrrrr|rrrrrr|r@{}}
\toprule
               &       &            &             & \multicolumn{6}{c|}{Baseline}                      & \multicolumn{6}{c|}{Candidate}                     &                   \\
Value          & $E_v$ & $\bar E_v$ & $S_{v,min}$ & $G_v$ & $R_v$ & $L_v$ & $N_v$ & $S_v$ & $\hat S_v$ & $G_v$ & $R_v$ & $L_v$ & $N_v$ & $S_v$ & $\hat S_v$ & $\Delta \hat S_v$ \\
\midrule
benevolence    & 373     & 627         & -686.5 & 261 & 318 & 55  & 366 & 341.0 & 0.61 & 418 & 361 & 12 & 209 & 662.5 & 0.80   & 0.19  \\
universalism   & 598     & 402         & -799.0 & 236 & 553 & 45  & 166 & 661.0 & 0.81 & 337 & 588 & 10 & 65  & 882.5 & 0.93   & 0.12  \\
self-direction & 310     & 690         & -655.0 & 175 & 192 & 118 & 515 & -8.5  & 0.39 & 550 & 292 & 18 & 140 & 754.0 & 0.85   & 0.46  \\
stimulation    & 531     & 469         & -765.5 & 184 & 473 & 58  & 285 & 456.5 & 0.69 & 301 & 508 & 23 & 168 & 702.0 & 0.83   & 0.14  \\
hedonism       & 408     & 592         & -704.0 & 160 & 308 & 100 & 432 & 152.0 & 0.50 & 452 & 395 & 13 & 140 & 764.0 & 0.86   & 0.36  \\
achievement    & 411     & 589         & -705.5 & 211 & 324 & 87  & 378 & 259.0 & 0.57 & 522 & 403 & 8  & 67  & 883.5 & 0.93   & 0.37  \\
power          & 453     & 547         & -726.5 & 237 & 379 & 74  & 310 & 387.0 & 0.64 & 427 & 429 & 24 & 120 & 772.0 & 0.87   & 0.22  \\
security       & 321     & 679         & -660.5 & 279 & 250 & 71  & 400 & 258.0 & 0.55 & 566 & 310 & 11 & 113 & 808.5 & 0.88   & 0.33  \\
conformity     & 249     & 751         & -624.5 & 232 & 180 & 69  & 519 & 83.5  & 0.44 & 313 & 201 & 48 & 438 & 247.0 & 0.54   & 0.10  \\
tradition      & 303     & 697         & -651.5 & 266 & 245 & 58  & 431 & 237.5 & 0.54 & 516 & 295 & 8  & 181 & 712.5 & 0.83   & 0.29  \\
\bottomrule
\end{tabular}

%% file: tab-results.tex
\begin{tabular}{@{}lp{3.5cm}@{}}
\toprule
Target LLM name       & Wizard-Vicuna-13B-Uncensored                                                                                          \\ 
Target LLM parameters & temperature: 0                                                                                                        \\
                      & max tokens: 256                                                                                                       \\
                      & prompt template and stop words: vicuna                                                                                \\
\midrule
Value theory          & Basic Human Values Theory                                                                                             \\
Value list            & benevolence, universalism, self-direction, stimulation, hedonism, achievement, power, security, conformity, tradition \\
Method name           & Valuesnet\_DeBERTa\_v3                                                                                                \\
Method parameters     & thresholds: assign label if result $\geq 0.5$                                                                         \\
                      & value extraction covers only the last two turns of the conversation                                                   \\
\midrule
Dataset name          & Commonsense-Dialogues                                                                                                 \\
Dataset type          & dialogues                                                                                                             \\
Dataset split         & 1K samples from train split                                                                                           \\
\midrule
Score p. baseline     & 0.57                                                                                                                  \\
Score p. candidate    & 0.83                                                                                                                  \\ 
\bottomrule
\end{tabular}